\documentclass[11pt]{article}

\usepackage[final]{acl}
\usepackage{times}
\usepackage{latexsym}
\usepackage{comment}
\usepackage{subcaption}
\usepackage[T1]{fontenc}
\usepackage{enumitem}

\usepackage[utf8]{inputenc}
\usepackage{newunicodechar}
\newunicodechar{ọ}{\d{o}}
\newunicodechar{ụ}{\d{u}}
\newunicodechar{ạ}{\d{a}}
\newunicodechar{ị}{\d{i}}
\newunicodechar{ẹ}{\d{e}}

\usepackage{microtype}
\usepackage{inconsolata}
\usepackage{amssymb}
\usepackage{graphicx}
\usepackage{tabularx,array,xltabular,booktabs}
\usepackage{tabularray}
\usepackage{caption}
\newcolumntype{Y}{>{\raggedright\arraybackslash}X}
\usepackage{listings}

\title{Robust Language Identification for Romansh Varieties}

\author{Charlotte Model \qquad Sina Ahmadi \qquad Jannis Vamvas \\
  University of Zurich\\
  \texttt{charlotte.model@uzh.ch, sina.ahmadi@uzh.ch, vamvas@cl.uzh.ch} \\
}

\begin{document}
\maketitle
\begin{abstract}
The Romansh language has several regional varieties, called \textit{idioms}, which sometimes have limited mutual intelligibility.
This linguistic diversity motivates the need for a language identification (LID) system that can distinguish between these idioms, yet to date there has been no well-documented effort to build one.
Since Romansh LID should also be able to recognize Rumantsch Grischun, a supra-regional variety that combines elements of several idioms, this makes for a novel and interesting classification problem.
In this paper, we present a LID system for Romansh idioms based on an SVM approach.
We evaluate our model on a newly curated benchmark across two domains and find that it reaches an average in-domain accuracy of 97\%, enabling applications such as idiom-aware spell checking or machine translation.
Our classifier is publicly available.\footnote{\url{https://github.com/ZurichNLP/romansh-lid}}
\end{abstract}

\section{Introduction}

Language identification (LID) is the task of automatically determining the language of text or speech. As a foundational component in natural language processing (NLP) pipelines, LID enables downstream applications such as machine translation, information retrieval, content moderation, and multilingual text processing by routing to the most appropriate language-specific system.
While state-of-the-art LID systems achieve high accuracy for widely-used languages~\citep{DBLP:conf/naacl/ChenADLA24}, distinguishing between closely-related language varieties remains a significant challenge~\cite{DBLP:conf/eacl/BurchellBTH24}. Unlike unrelated languages that exhibit clear phonological, morphological, and lexical differences, closely-related varieties have fewer unique linguistic features, making them difficult to differentiate automatically.
The challenges of LID are compounded for varieties spoken in limited geographic regions, where the scarcity of diverse digital data hinders the development and generalization of robust LID systems.

In this paper, we present a LID system for Romansh, a collection of closely-related linguistic varieties native to Switzerland and spoken by approximately 40,000 people \citep{gruenert}. Due to Romansh's low-resource status, no well-documented effort in the NLP community has developed a LID system capable of distinguishing between its varieties; existing multilingual LID systems either ignore it entirely or treat it as a monolithic language. Our contributions are threefold:

\begin{enumerate}[label=(\arabic*)]
\item We curate a novel benchmark dataset spanning multiple domains (news, broadcast transcripts, textbooks, and newsroom notes) to enable systematic evaluation of idiom classification;
\item We develop an SVM-based classifier that achieves 97\% average in-domain accuracy across all six varieties; and
\item We provide qualitative analysis on the LID task for Romansh idioms.
\end{enumerate}
Our work demonstrates that effective LID for closely-related low-resource varieties is achievable with carefully designed features and appropriate training data, providing a model for similar efforts in other endangered and regional language contexts.  

\section{Romansh and its Varieties}

The term \textit{Romansh} refers to a collection of closely related linguistic varieties of Rhaetian descent native to the canton of Grisons in Switzerland. These varieties, known as \textit{idioms}, comprise five historically distinct forms: \textit{Sursilvan}, \textit{Sutsilvan}, \textit{Surmiran}, \textit{Puter}, and \textit{Vallader}. To facilitate official communication with the government, a sixth, artificially-created variety was introduced in 1982 to act as the established written standard, known as \textit{Rumantsch Grischun} (RG) \citep{gruenert}. Communities speaking different idioms are distributed within an area of approximately 7,000 km\textsuperscript{2} across Grisons.

The geographical proximity of certain idioms is reflected in their linguistic similarity. According to Lia Rumantscha, the organization tasked by the Swiss government to represent and preserve the Romansh identity, the two idioms spoken in the Rhine valley, Sursilvan and Sutsilvan, are lexically closely related, as are Puter and Vallader, spoken in the upper and lower Engadine valley respectively \citep{lia-rumantscha-facts}. Surmiran, spoken in the central region, often acts as a bridge between these two idiom groups. Finally, since RG was developed on the basis of Sursilvan, Vallader, and Surmiran, it shares features with multiple idioms \citep{Anderson2016}.

\section{Related Work}
\subsection{Language Identification}
LID has been widely studied in NLP, with growing importance for filtering the multilingual corpora used to train language models~\cite{foroutan2025conlid}. When applied to closely-related varieties and dialects, especially within low-resource languages, the task poses significant challenges due to the limited availability of linguistic resources and annotated datasets. Research highlights the difficulty of accurately distinguishing among variants, as demonstrated in studies focused on dialectal Arabic~\cite{10.1145/3747290} and Indo-Aryan languages such as Bhojpuri and Assamese~\cite{10.1145/3458250}. Recent techniques explore machine learning and deep learning architectures, leveraging cross-lingual transfer learning and community-driven approaches that emphasize regional linguistic families~\cite{10.18653/v1/2021.sigtyp-1.11}.

\subsection{Romansh NLP}

Ongoing projects at Lia Rumantscha for Romansh NLP focus on offering digital services to translate from Romansh into German, the most widespread official language of Switzerland, and vice versa. These services include idiom-specific dictionaries hosted across several portals: Pledari Grond covers Surmiran, Sutsilvan, RG and most recently Sursilvan directly in its interface\footnote{\url{https://www.pledarigrond.ch}}, the Uniun dals Grischs\footnote{\url{https://www.udg.ch}} serves Puter and Vallader, and the Dicziunari Rumantsch is an app that aggregates this data on mobile\footnote{\url{https://www.dicziunari.ch}}. Pledari Grond also provides a spell-checker UI for RG\footnote{\url{https://www.pledarigrond.ch/rumantschgrischun/spellchecker}}. One property shared by all these projects is that the user must select the desired idiom before translation or spell-checking can take place, limiting their usability to cases where the user already knows what idiom a text is in.

More recently, resources and models have been released that specifically focus on Romansh, including the Mediomatix Corpus~\cite{hopton2025mediomatix}, containing parallel sentences extracted from schoolbooks, and WMT24++ \cite{vamvas-et-al-2025-expanding}, containing Romansh translations of the WMT24++ benchmark in machine translation. Additionally, the Swiss-made LLM Apertus has specifically incorporated Romansh data for post-training~\cite[p.~94]{apertus2025}.

\section{Data}
\label{sec:data}

For the training and evaluation of Romansh LID systems, we compile five sources of textual data spanning dictionary entries, journalistic articles, broadcast transcripts, newsroom notes, and school textbooks:

\begin{itemize}[noitemsep]
    \item \textbf{Pledari Grond} (PG), a comprehensive Romansh-German dictionary covering all Romansh idioms. 
    \item \textbf{La Quotidiana} (LQ), a Romansh newspaper with daily idiom-annotated content. We use WordPress dumps from 2021 to 2025. 
    \item \textbf{Radiotelevisiun Svizra Rumantscha} (RTR), validated speech transcripts from Romansh broadcasts, annotated by idiom.
    \item \textbf{RTR Telesguard Notes} (TG), pre-broadcast notes written by journalists in their native idioms (excluding RG). 
    \item \textbf{Mediomatix Textbooks} (TB), parallel scholastic material per idiom (excluding RG), recently released by \citet{hopton2025mediomatix}. 
\end{itemize}

\subsection{Preprocessing}
For each source we extract the idiom label and the main text fields. We then apply minimal cleaning to all samples by i) removing intra-class duplicates (exact string match within idiom); ii) stripping HTML/markup and collapsing repeated whitespace and newlines; iii) dropping empty/\texttt{None}/non-letter-only items. Additionally, we remove source artifacts that do not carry language cues such as dictionary markers, e.g., sense numerals, ``cf.'' stubs, editorial signatures, worksheet placeholders and long underscore sequences. RTR content showed no recurring artifacts. Noisy or non-Rumantsch snippets were discarded and clear mislabels, if noticed visually, were manually corrected by dictionary checks.

Appendix~\ref{app:data_sources} summarises the export statistics for each data source after this cleaning, and provides links to the publicly available datasets.

Following \citet{bernier-2023}, we run exact and (where feasible) near-duplicate detection across all the data sources after merging and before splitting them. Exact duplicates judged valid in multiple idioms are kept but routed to the training split.

\subsection{Named-Entity Masking}
To reduce reliance on lexical memorisation, we produce two training variants: \texttt{masked} and \texttt{unmasked}. For \texttt{masked}, we run a fine-tuned named entity recognition model based on SwissBERT~\cite{vamvas-etal-2023-swissbert} (zero-shot on RG) with conservative heuristics (min length, standalone tokens, score $\ge 0.98$) and replace matched spans with \texttt{\$NE\$}. Pledari Grond is left unmasked due to very few named entities and high inference cost. Because masking replaces spans in place rather than dropping samples, both variants contain the same 487{,}172 samples and differ only in token count (12.43M unmasked vs.\ 12.38M masked).

\section{Experimental Setup}

\subsection{Data Splits}
Following these steps, we finally split all the datasets into train, validation and test sets. For the latter, we use multiple test sets to disentangle domain, balance, and comparability effects.

\begin{itemize}
    \item \textbf{Train} (\texttt{train-set}): PG + LQ + RTR + TB (unbalanced). Both masked and unmasked variants used in training experiments.
    \item \textbf{Dev} (\texttt{dev-set}): Balanced, in-domain RTR, \textbf{6{,}000} samples (1k/idiom; avg.\ 45.9 tokens).
    \item \textbf{Test-A} (\texttt{test-a}): In-domain, \emph{unbalanced} LQ, \textbf{6{,}000} samples (avg.\ 528.5 tokens).
    \item \textbf{Test-B} (\texttt{test-b}): In-domain, \emph{balanced} RTR, \textbf{6{,}000} samples (avg.\ 45.7 tokens).
    \item \textbf{Test-C} (\texttt{test-c}): In-domain, approximately \emph{balanced} TB (no RG), \textbf{6{,}000} samples (avg.\ 85.8 tokens).
    \item \textbf{Test-D} (\texttt{test-d}): \emph{Out-of-domain} TG (no RG), \textbf{9{,}607} samples (avg.\ 151.9 tokens).
\end{itemize}

To reflect realistic LID inputs, we did not lowercase or strip punctuation on dev/test; we only removed non-letter-only items and ensured non-empty (post-strip) text/labels.

\begin{table*}[t]
    \centering
    \begin{tabular}{@{}lrrrr@{}}
    \toprule
    Classifier & Accuracy & Macro F1 & Weighted F1 & Macro Recall \\
    \midrule
    Logistic Regression (LR) & 79.5 & 76.9 & 79.1 & 74.6 \\
    Linear SVM & 78.1 & 75.4 & 77.9 & 73.9 \\
    SGD (Linear SVM) & 76.3 & 72.4 & 75.2 & 68.0 \\
    Naive Bayes (counts) & 75.1 & 72.1 & 74.6 & 70.6 \\
    Naive Bayes (TF) & 75.1 & 72.1 & 74.6 & 70.6 \\
    Naive Bayes (TF-IDF) & 73.7 & 72.0 & 73.9 & 71.7 \\
    SGD (Logistic Regression) & 73.5 & 69.0 & 72.0 & 63.6 \\
    Majority Baseline & 35.1 & 8.7 & 18.3 & 16.7 \\
    \bottomrule
    \end{tabular}
    \caption{Results from the preliminary experiments with different baseline classifiers, where classifiers are ordered by macro F1 score.}
    \label{tab:prelim_results}
\end{table*}

\subsection{Classification}
We frame our task as a supervised multi-class classification problem where the task is to assign each text sample to one of six varieties.

We extract bag-of-$n$-grams features combining word unigrams and overlapping character $n$-grams ($n \in \{1,2,3,4\}$), represented as TF--IDF vectors with sublinear term-frequency scaling and $\ell_2$ normalisation. We compare four families of linear classifiers, all from \texttt{scikit-learn} \citep{scikit-learn}, each with explicit $\ell_2$ regularisation by default: 
(i) Logistic Regression (\texttt{LogisticRegression}, multinomial, $\ell_2$ penalty, inverse regularisation strength $C$, \texttt{saga} solver, 5{,}000 max iterations); 
(ii) linear SVM (\texttt{LinearSVC}, squared-hinge loss, $\ell_2$ penalty, \texttt{dual=False}, inverse regularisation strength $C$); 
(iii) two stochastic-gradient variants (\texttt{SGDClassifier})—one with hinge loss matching the SVM objective (SGD-SVM), the other with log loss matching the LR objective (SGD-LR), both with $\ell_2$ regularisation strength $\alpha = 10^{-4}$, 5{,}000 max iterations, and early stopping; 
(iv) Multinomial and Complement Na\"{\i}ve Bayes baselines (additive smoothing $\alpha = 1.0$). 

Baseline values were the \texttt{scikit-learn} defaults ($C = 1.0$ for SVM and LR), and were subsequently tuned by randomised search (\S\,5.3). The two SGD variants share the SVM and LR loss functions respectively, but differ in optimiser: \texttt{LogisticRegression} uses the full-batch \texttt{saga} solver, whereas \texttt{SGDClassifier} uses mini-batch SGD with a fixed learning-rate schedule. This isolates optimiser effects from loss-function effects.

\subsection{Hyperparameter Optimization}
We perform 40-iteration randomized searches over pipeline hyperparameters using stratified 5-fold cross-validation on a stratified 20\% subset of the training data, optimising for macro~$F_1$ (\texttt{RandomizedSearchCV}, \texttt{random\_state}~=~42 throughout). The search space included the regularisation parameters of each classifier ($C$ drawn log-uniformly from $[10^{-2}, 4]$ for SVM and $[10^{-2}, 2]$ for LR), the penalty norm ($\ell_2$, $\ell_1$, elastic net for LR; $\ell_2$ and $\ell_1$ for SVM), the elastic-net mixing ratio $l_1\_ratio \in [0.05, 0.9]$, the SVM loss (hinge / squared-hinge), the character $n$-gram range ($(1,3)$ or $(1,4)$), the word $n$-gram range ($(1,1)$ or $(1,2)$), and the minimum document frequency (1 or 2). Final models were trained on both unmasked and masked variants of the training set.

The best LR configuration (penalty~=~$\ell_1$, $C \approx 1.94$, $l_1\_ratio \approx 0.47$, character $(1,4)$-grams, word $(1,2)$-grams) reached 96.8 macro~$F_1$ on the dev set, while the best SVM configuration (penalty~=~$\ell_1$, $C \approx 0.62$, squared-hinge loss, character $(1,4)$-grams with $\mathrm{min\_df} = 2$, word unigrams only) reached 97.1 macro~$F_1$ on the dev set. 

\section{Results}
\subsection{Overall Results}

Table~\ref{tab:prelim_results} reports the preliminary baseline comparison. All classifiers substantially outperformed the majority-class baseline, with macro $F_1$ scores clustering within 8 points of each other. LR achieved the highest score (76.9), followed closely by linear SVM (75.4), suggesting that for this task feature representation matters more than classifier choice. We retained LR and SVM for hyperparameter optimisation.

After tuning (\S\,5.3), the ordering reversed: SVM (97.1 dev macro $F_1$) slightly outperformed LR (96.8). We therefore selected SVM for the remaining experiments.

\begin{figure*}[ht]
    \centering
    \begin{subfigure}[b]{0.48\textwidth}
        \centering
        \includegraphics[width=\linewidth]{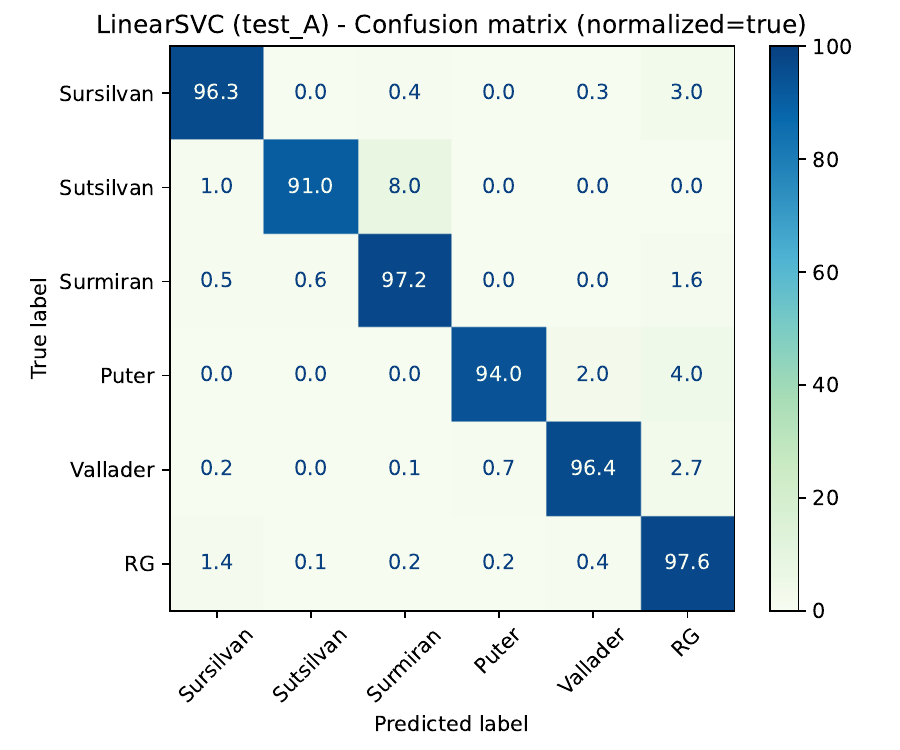}
        \caption{\texttt{test-a}}
        \label{fig:svm-masked-train-test-a-conf}
    \end{subfigure}
    \hfill
    \begin{subfigure}[b]{0.48\textwidth}
        \centering
        \includegraphics[width=\linewidth]{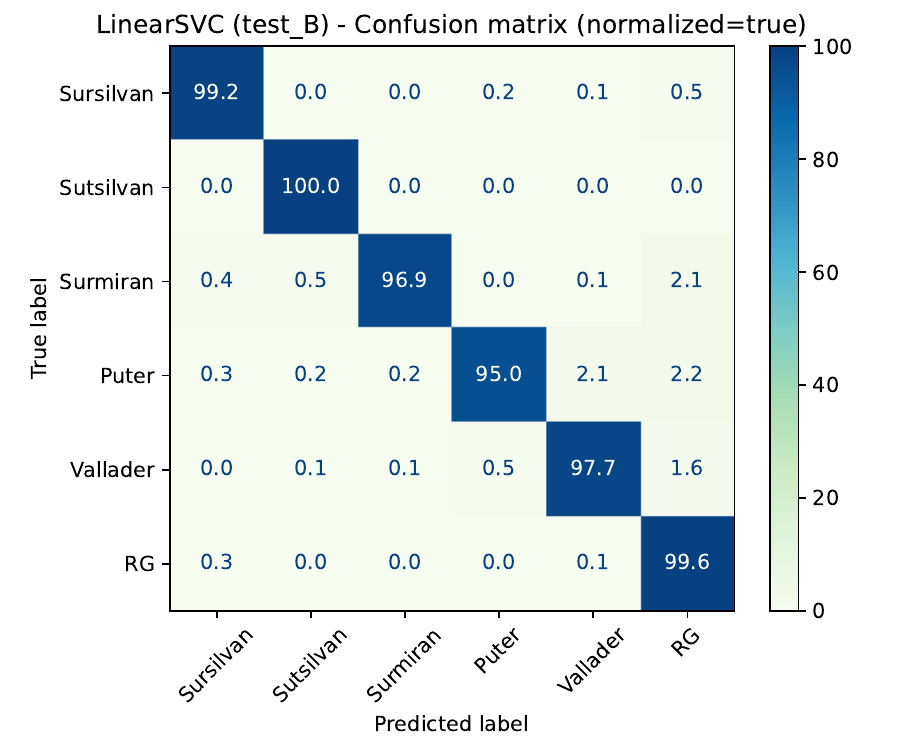}
        \caption{\texttt{test-b}}
        \label{fig:svm-masked-train-test-b-conf}
    \end{subfigure}
    
    \vspace{1em}
    
    \begin{subfigure}[b]{0.48\textwidth}
        \centering
        \includegraphics[width=\linewidth]{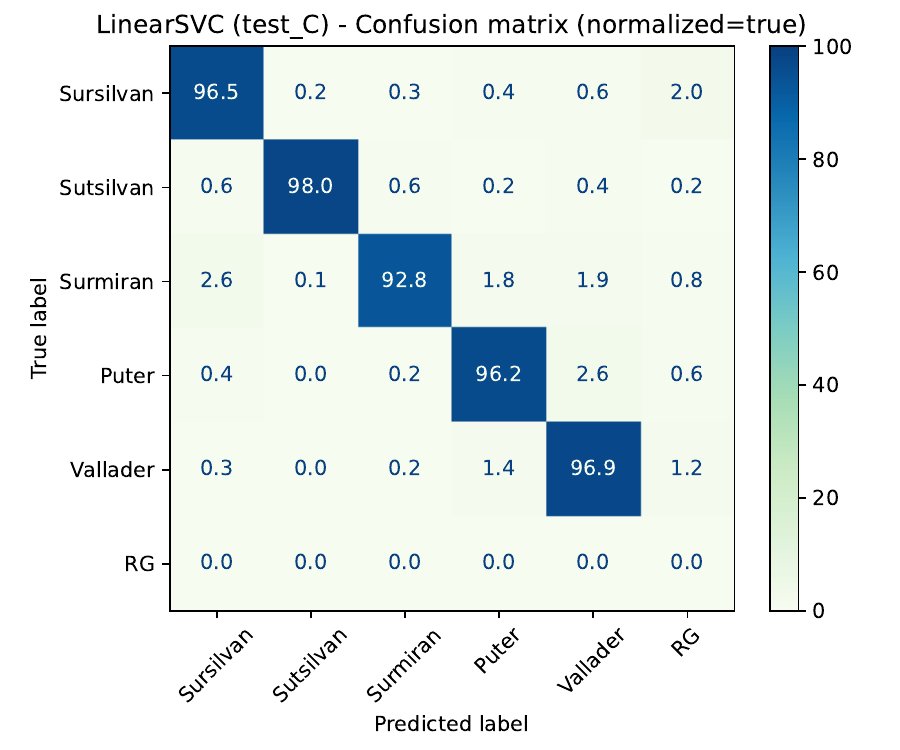}
        \caption{\texttt{test-c}}
        \label{fig:svm-masked-train-test-c-conf}
    \end{subfigure}
    \hfill
    \begin{subfigure}[b]{0.48\textwidth}
        \centering
        \includegraphics[width=\linewidth]{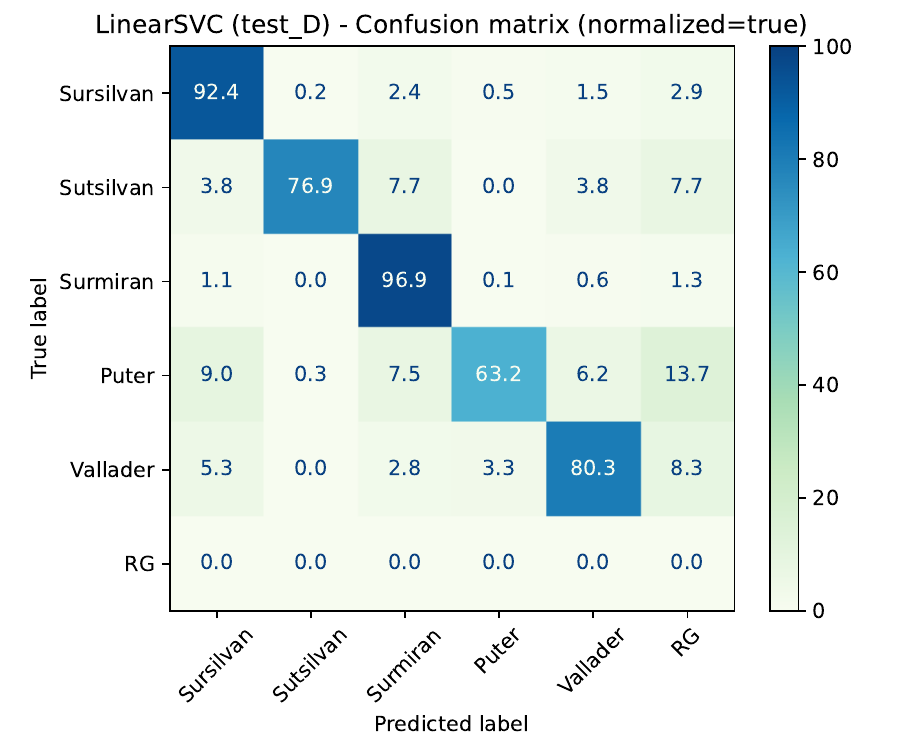}
        \caption{\texttt{test-d}}
        \label{fig:svm-masked-train-test-d-conf}
    \end{subfigure}
    
    \caption{Row-normalized confusion matrices for all test sets. The model achieves near-perfect classification on the balanced in-domain set (\texttt{test-b}), while confusion increases on out-of-domain data (\texttt{test-d}).}
    \label{fig:svm-masked-train-conf-matrices}
\end{figure*}

Table \ref{tab:svm-train-and-dev} presents the performance of our optimized linear SVM classifier across the four different test sets. The model achieves macro F1 scores ranging from 98.1 on the balanced in-domain test set (\texttt{test-b}) to 69.1 on the unbalanced out-of-domain set (\texttt{test-d}), demonstrating this classical machine learning approach is effective for Romansh LID under favorable conditions while struggling with out-of-domain and noisy data. These results compare favorably to similar language discrimination tasks. Recent VarDial shared tasks on Italian dialects and French varieties report best macro F1 scores of 74.6 and 34.4 respectively, while the Dravidian Language Identification task achieved a score of 93, closer to our in-domain results \cite{vardial-2021, vardial-2022}.

\begin{table}[t]
    \centering
    \begin{tblr}{
      colspec = {@{}lllrr@{}},
      hline{1,6} = {-}{0.08em},
      hline{2} = {-}{},
    }
    Test Set  & Domain & Balance & Acc. & F1 \\
    \texttt{test-a}    & in-domain &  unbal.  & 96.8 & 94.7 \\
    \texttt{test-b}    & in-domain &  bal.    & 98.1 & 98.1 \\
    \texttt{test-c}    & in-domain &  bal.    & 96.2 & 80.5 \\
    \texttt{test-d}    & out-domain &  unbal.  & 90.7 & 69.1
    \end{tblr}
    \caption{Accuracy and macro F1 scores per test set. The highest scores are achieved on the balanced (bal.) vs. unbalanced (unbal.), in-domain set \texttt{test-b}.}
    \label{tab:svm-train-and-dev}
\end{table}

\begin{table*}[!t]
\centering
\begin{tblr}{
  colspec = {@{}lrrrr@{}},
  hline{1,7} = {-}{0.08em},
  hline{3} = {-}{},
  cell{1}{2} = {c=2}{c},
  cell{1}{4} = {c=2}{c},
  cell{1}{1} = {r=2}{},
}
Test Set & Accuracy & & Macro F1 & \\
         & Masked & Unmasked & Masked & Unmasked \\
\texttt{test-a}   & 96.8 & 96.9 & 94.7 & 94.5 \\
\texttt{test-b}   & 98.1 & 98.8 & 98.1 & 98.8 \\
\texttt{test-c}   & 96.2 & 96.2 & 80.5 & 80.4 \\
\texttt{test-d}   & 90.7 & 90.5 & 69.1 & 69.0 \\
\end{tblr}
\caption{Comparison of masked vs.\ unmasked training. No major effect can be observed.}
\label{tab:masked_vs_unmasked}
\end{table*}

\subsection{Per Idiom Performances}
Figure~\ref{fig:svm-masked-train-conf-matrices} shows confusion matrices for each test set. The model achieves consistently high recall across idioms on \texttt{test-b} (95.0--100.0), with perfect classification of Sutsilvan samples. Performance varies more on other sets: \texttt{test-a} shows recalls of 91.0--97.6, \texttt{test-c} shows 92.8--98.0, while \texttt{test-d} exhibits the widest range (63.2--96.9). The notably low recall for Puter in \texttt{test-d} (63.2) warrants investigation, as samples were frequently misclassified as RG. This may reflect the informal, potentially noisy nature of the Telesguard notes data. Our analysis of misclassifications reveals several data quality issues. In \texttt{test-a}, many errors involved samples containing primarily named entities or German text---artifacts from the journalism data source. In \texttt{test-c}, misclassifications often occurred on noisy samples containing mainly numbers, punctuation, or very short texts from the textbook data.

\begin{table}[ht]
\centering
\small
\resizebox{\linewidth}{!}{
\begin{tblr}{
  colspec = {@{}l cc cc cc@{}},
  hline{1,8} = {-}{0.08em},
  hline{2} = {-}{},
}
Idiom    & 1st & Type & 2nd & Type & 3rd & Type \\
Sursilvan  & ei  & char & scha & char & iu & char \\       
Sutsilvan  & àn  & char & ù    & char & eing & char \\    
Surmiran & eir & char & dall & char & ous & char \\
Puter    & ron & char & aunt & char & ọ & char \\
Vallader & ạ   & char & ọn   & char & ị & char \\
RG       & \textvisiblespace & char & vegnis & word & ì & char \\          
\end{tblr}
}
\caption{Top 3 most discriminative features per idiom. ``char'' denotes character $n$-gram and ``word'' denotes word unigram. Character $n$-grams emerge as the most discriminative features across all varieties, with empty space \textvisiblespace~being the most informative one for RG.}
\label{tab:discrim_feats}
\end{table}

\subsection{Discriminative Features}
The most informative features are predominantly character $n$-grams rather than word unigrams, aligning with findings from similar LID tasks. We present these features in Table~\ref{tab:discrim_feats}. Several patterns emerge: notably, Puter and Vallader features include underdotted characters (ọ, ụ, ạ, ị, ẹ) that are unique to these idioms in the Pledari Grond data. Although these features are highly discriminative for our classifier, they are not part of the standard orthography of Puter and Vallader: in Pledari Grond they function as phonetic stress markers, and would not appear in running text written in either idiom outside a dictionary setting. The whitespace character appearing as the top feature for RG is an artefact of severe class imbalance in the training data ($\approx$171K RG samples versus 44K--86K for other idioms; see Appendix~\ref{app:data_stats}), since whitespace sequences are normalised to a single space during preprocessing and RG samples then dominate single-space $n$-grams.

We tested two ablations to address these artefacts: (i) restricting character $n$-grams to within-word boundaries only (rather than allowing them to span across whitespace), and (ii) removing character $1$-grams from the feature set entirely. Both decreased overall macro~$F_1$, most plausibly because they remove the underdotted characters and other diacritics that supply most of the discriminative signal for Puter and Vallader.

Future work should explore balancing the training data, which we expect to be a more robust remedy for the whitespace artefact than feature-side filtering.

\subsection{Impact of Named Entities}
Comparing models trained on masked vs.\ unmasked data, as presented in Table~\ref{tab:masked_vs_unmasked}, shows negligible performance differences ($\leq$ 0.1 F1 points on most sets; on \texttt{test-b} the unmasked variant is 0.7 F1 higher, the largest gap). This confirms that named entities provide no idiom-specific information---unsurprising given the geographic proximity of Romansh communities. News outlets like La Quotidiana and RTR report region-wide events in multiple idioms, textbooks cover similar subjects across idioms, and RG serves as a pan-idiom standard.

\section{Conclusion}

We presented the first documented language identification system for Romansh idioms, capable of distinguishing between the five historical varieties (Sursilvan, Sutsilvan, Surmiran, Puter, and Vallader) and the standardised RG. Using an SVM classifier with character and word n-gram features, our model achieves up to 98\% macro F1 on balanced in-domain data. Our experiments confirm that character n-grams are the most discriminative features for this task, with idiom-specific diacritics and orthographic patterns providing strong classification signals. Furthermore, we found that named entity masking has negligible impact on performance, suggesting that the classifier relies on genuine linguistic features rather than memorizing location-specific proper nouns.

\section*{Limitations}

Our work has several limitations. First, the training data exhibits substantial class imbalance, with RG overrepresented due to the Pledari Grond dictionary data, which may explain artifacts such as whitespace emerging as a top discriminative feature. Second, performance degrades on out-of-domain data (\texttt{test-d}), where macro F1 drops to 69.1, indicating limited generalization to informal text genres like newsroom notes. Third, some discriminative features, particularly the underdotted characters unique to Pledari Grond, may not generalize to texts following standard orthographic conventions. Finally, our evaluation is restricted to written text; spoken language identification for Romansh remains unexplored. Despite these limitations, our system provides a practical tool for downstream applications such as idiom-aware spell checking and machine translation routing, and establishes a baseline for future work on Romansh and other low-resource regional language varieties.

\section*{Acknowledgments}
This work is based on a Bachelor's thesis that was presented to the University of Zurich.\footnote{\url{https://seafile.ifi.uzh.ch/f/96df2a17539546e7a192/}}
We thank Lia Rumantscha and RTR for their support and for facilitating access to the data sources used in this study, including Pledari Grond, La Quotidiana, and RTR materials.
We also thank Uniun dals Grischs for making dictionary data for Puter and Vallader available to us for research use, and Ignacio Pérez Prat for helpful feedback.
Their commitment to preserving and promoting the Romansh language made this research possible.

\bibliography{custom}

@incollection{gruenert,
  author    = {Gr{\"u}nert, Matthias},
  title     = {R{\"a}toromanisch},
  booktitle = {Sprachenr{\"a}ume der Schweiz. Band 1: Sprachen},
  editor    = {Glaser, Elvira and Kabatek, Johannes and Sonnenhauser, Barbara},
  publisher = {Narr Francke Attempto},
  address   = {T{\"u}bingen},
  year      = {2024},
  month     = nov,
  pages     = {156--184},
  isbn      = {978-3-381-10401-7},
  doi       = {10.24053/9783381104024} 
}

@inproceedings{DBLP:conf/naacl/ChenADLA24,
  author       = {Wei{-}Rui Chen and
                  Ife Adebara and
                  Khai Duy Doan and
                  Qisheng Liao and
                  Muhammad Abdul{-}Mageed},
  editor       = {Kevin Duh and
                  Helena G{\'{o}}mez{-}Adorno and
                  Steven Bethard},
  title        = {Fumbling in Babel: An Investigation into ChatGPT's Language Identification
                  Ability},
  booktitle    = {Findings of the Association for Computational Linguistics: {NAACL}
                  2024, Mexico City, Mexico, June 16-21, 2024},
  pages        = {4387--4413},
  publisher    = {Association for Computational Linguistics},
  year         = {2024},
  url          = {https://doi.org/10.18653/v1/2024.findings-naacl.274},
  doi          = {10.18653/V1/2024.FINDINGS-NAACL.274},
  bibsource    = {dblp computer science bibliography, https://dblp.org}
}

@inproceedings{DBLP:conf/eacl/BurchellBTH24,
  author       = {Laurie Burchell and
                  Alexandra Birch and
                  Robert P. Thompson and
                  Kenneth Heafield},
  editor       = {Yvette Graham and
                  Matthew Purver},
  title        = {Code-Switched Language Identification is Harder Than You Think},
  booktitle    = {Proceedings of the 18th Conference of the European Chapter of the
                  Association for Computational Linguistics, {EACL} 2024 - Volume 1:
                  Long Papers, St. Julian's, Malta, March 17-22, 2024},
  pages        = {646--658},
  publisher    = {Association for Computational Linguistics},
  year         = {2024},
  url          = {https://aclanthology.org/2024.eacl-long.38},
  bibsource    = {dblp computer science bibliography, https://dblp.org}
}

@inproceedings{vamvas-et-al-2025-expanding,
    title = "Expanding the {WMT}24++ Benchmark with {R}umantsch {G}rischun, {S}ursilvan, {S}utsilvan, {S}urmiran, {P}uter, and {V}allader",
    author = "Vamvas, Jannis  and
      P{\'e}rez Prat, Ignacio  and
      Soliva, Not  and
      Baltermia-Guetg, Sandra  and
      Beeli, Andrina  and
      Beeli, Simona  and
      Capeder, Madlaina  and
      Decurtins, Laura  and
      Gregori, Gian Peder  and
      Hobi, Flavia  and
      Holderegger, Gabriela  and
      Lazzarini, Arina  and
      Lazzarini, Viviana  and
      Rosselli, Walter  and
      Vital, Bettina  and
      Rutkiewicz, Anna  and
      Sennrich, Rico",
    editor = "Haddow, Barry  and
      Kocmi, Tom  and
      Koehn, Philipp  and
      Monz, Christof",
    booktitle = "Proceedings of the Tenth Conference on Machine Translation",
    month = nov,
    year = "2025",
    address = "Suzhou, China",
    publisher = "Association for Computational Linguistics",
    url = "https://aclanthology.org/2025.wmt-1.79/",
    pages = "1028--1047",
    ISBN = "979-8-89176-341-8"
}

@inproceedings{bernier-2023,
  address =       {Dubrovnik, Croatia},
  author =        {Bernier-colborne, Gabriel and Goutte, Cyril and
                   Leger, Serge},
  booktitle =     {Tenth Workshop on NLP for Similar Languages,
                   Varieties and Dialects (VarDial 2023)},
  editor =        {Scherrer, Yves and Jauhiainen, Tommi and
                   Ljube{\v{s}}i{\'c}, Nikola and Nakov, Preslav and
                   Tiedemann, J{\"o}rg and Zampieri, Marcos},
  month =         may,
  pages =         {142--151},
  publisher =     {Association for Computational Linguistics},
  title =         {Dialect and Variant Identification as a Multi-Label
                   Classification Task: A Proposal Based on
                   Near-Duplicate Analysis},
  year =          {2023},
  doi =           {10.18653/v1/2023.vardial-1.15},
  url =           {https://aclanthology.org/2023.vardial-1.15/},
}

@inproceedings{vamvas-etal-2023-swissbert,
    title = "{S}wiss{BERT}: The Multilingual Language Model for {S}witzerland",
    author = {Vamvas, Jannis  and
      Gra{\"e}n, Johannes  and
      Sennrich, Rico},
    editor = {Ghorbel, Hatem  and
      Sokhn, Maria  and
      Cieliebak, Mark  and
      H{\"u}rlimann, Manuela  and
      de Salis, Emmanuel  and
      Guerne, Jonathan},
    booktitle = "Proceedings of the 8th edition of the Swiss Text Analytics Conference",
    month = jun,
    year = "2023",
    address = "Neuchatel, Switzerland",
    publisher = "Association for Computational Linguistics",
    url = "https://aclanthology.org/2023.swisstext-1.6/",
    pages = "54--69"
}

@article{scikit-learn,
  author =        {Pedregosa, F. and Varoquaux, G. and Gramfort, A. and
                   Michel, V. and Thirion, B. and Grisel, O. and
                   Blondel, M. and Prettenhofer, P. and Weiss, R. and
                   Dubourg, V. and Vanderplas, J. and Passos, A. and
                   Cournapeau, D. and Brucher, M. and Perrot, M. and
                   Duchesnay, E.},
  journal =       {Journal of Machine Learning Research},
  pages =         {2825--2830},
  title =         {Scikit-learn: Machine Learning in {P}ython},
  volume =        {12},
  year =          {2011},
}

@article{10.1145/3747290,
    author = "Dahou, Abdelghani and Dahou, Abdelhalim Hafedh and Chéragui, Mohamed Amine and Abdedaiem, Amin and A. Al‐qaness, Mohammed A. and Elaziz, Mohamed Abd and Ewees, Ahmed A. and Zheng, Zhonglong",
    doi = "10.1145/3747290",
    title = "A Survey on Dialect Arabic Processing and Analysis: Recent Advances and Future Trends",
    journal = "Acm Transactions on Asian and Low-Resource Language Information Processing",
    year = "2025"
}

@article{10.1145/3458250,
    author = "Mundotiya, Rajesh Kumar and Singh, Manish Kumar and Kapur, Rahul and Mishra, Swasti and Singh, Anil Kumar",
    doi = "10.1145/3458250",
    title = "Linguistic Resources for Bhojpuri, Magahi, and Maithili: Statistics About Them, Their Similarity Estimates, and Baselines for Three Applications",
    journal = "Acm Transactions on Asian and Low-Resource Language Information Processing",
    year = "2021"
}

@inproceedings{10.18653/v1/2021.sigtyp-1.11,
    title = "{SIGTYP} 2021 Shared Task: Robust Spoken Language Identification",
    author = "Salesky, Elizabeth  and
      Abdullah, Badr M.  and
      Mielke, Sabrina  and
      Klyachko, Elena  and
      Serikov, Oleg  and
      Ponti, Edoardo Maria  and
      Kumar, Ritesh  and
      Cotterell, Ryan  and
      Vylomova, Ekaterina",
    editor = {Vylomova, Ekaterina  and
      Salesky, Elizabeth  and
      Mielke, Sabrina  and
      Lapesa, Gabriella  and
      Kumar, Ritesh  and
      Hammarstr{\"o}m, Harald  and
      Vuli{\'c}, Ivan  and
      Korhonen, Anna  and
      Reichart, Roi  and
      Ponti, Edoardo Maria  and
      Cotterell, Ryan},
    booktitle = "Proceedings of the Third Workshop on Computational Typology and Multilingual NLP",
    month = jun,
    year = "2021",
    address = "Online",
    publisher = "Association for Computational Linguistics",
    url = "https://aclanthology.org/2021.sigtyp-1.11/",
    doi = "10.18653/v1/2021.sigtyp-1.11",
    pages = "122--129"
}

@inproceedings{foroutan2025conlid,
    title = "{C}on{LID}: Supervised Contrastive Learning for Low-Resource Language Identification",
    author = "Foroutan, Negar  and
      Saydaliev, Jakhongir  and
      Kim, Grace  and
      Bosselut, Antoine",
    editor = "Demberg, Vera  and
      Inui, Kentaro  and
      Marquez, Llu{\'i}s",
    booktitle = "Proceedings of the 19th Conference of the {E}uropean Chapter of the {A}ssociation for {C}omputational {L}inguistics (Volume 1: Long Papers)",
    month = mar,
    year = "2026",
    address = "Rabat, Morocco",
    publisher = "Association for Computational Linguistics",
    url = "https://aclanthology.org/2026.eacl-long.315/",
    doi = "10.18653/v1/2026.eacl-long.315",
    pages = "6693--6708",
    ISBN = "979-8-89176-380-7"
}

@incollection{Anderson2016,
  address =       {Oxford},
  author =        {Anderson, Stephen R.},
  booktitle =     {The Oxford Guide to the Romance Languages},
  editor =        {Ledgeway, Adam and Maiden, Martin},
  month =         jun,
  pages =         {169--184},
  publisher =     {Oxford University Press},
  title =         {Romansh (Rumantsch)},
  year =          {2016},
  doi =           {10.1093/acprof:oso/9780199677108.003.0012},
  isbn =          {9780199677108},
  url =           {https://doi.org/10.1093/acprof:oso/9780199677108.003.0012},
}

@misc{lia-rumantscha-facts,
  author =        {{Lia Rumantscha}},
  howpublished =
  {\url{https://www.liarumantscha.ch/sites/default/files/2023-07/PDF%20cumplet_d.pdf}},
  note =          {Accessed: 2025-06-17},
  title =         {{Facts}},
  year =          {2015},
}

@inproceedings{hopton2025mediomatix,
    title = "The Mediomatix Corpus: Parallel Data for {R}omansh Language Varieties via Comparable Schoolbooks",
    author = {Hopton, Zachary  and
      Vamvas, Jannis  and
      B{\"u}chler, Andrin  and
      Rutkiewicz, Anna  and
      Cathomas, Rico  and
      Sennrich, Rico},
    editor = "Demberg, Vera  and
      Inui, Kentaro  and
      Marquez, Llu{\'i}s",
    booktitle = "Findings of the {A}ssociation for {C}omputational {L}inguistics: {EACL} 2026",
    month = mar,
    year = "2026",
    address = "Rabat, Morocco",
    publisher = "Association for Computational Linguistics",
    url = "https://aclanthology.org/2026.findings-eacl.16/",
    doi = "10.18653/v1/2026.findings-eacl.16",
    pages = "290--306",
    ISBN = "979-8-89176-386-9"
}

@misc{apertus2025,
      title={Apertus: Democratizing Open and Compliant LLMs for Global Language Environments}, 
      author={Alejandro Hernández-Cano and Alexander Hägele and Allen Hao Huang and Angelika Romanou and Antoni-Joan Solergibert and Barna Pasztor and Bettina Messmer and Dhia Garbaya and Eduard Frank Ďurech and Ido Hakimi and Juan García Giraldo and Mete Ismayilzada and Negar Foroutan and Skander Moalla and Tiancheng Chen and Vinko Sabolčec and Yixuan Xu and Michael Aerni and Badr AlKhamissi and Ines Altemir Marinas and Mohammad Hossein Amani and Matin Ansaripour and Ilia Badanin and Harold Benoit and Emanuela Boros and Nicholas Browning and Fabian Bösch and Maximilian Böther and Niklas Canova and Camille Challier and Clement Charmillot and Jonathan Coles and Jan Deriu and Arnout Devos and Lukas Drescher and Daniil Dzenhaliou and Maud Ehrmann and Dongyang Fan and Simin Fan and Silin Gao and Miguel Gila and María Grandury and Diba Hashemi and Alexander Hoyle and Jiaming Jiang and Mark Klein and Andrei Kucharavy and Anastasiia Kucherenko and Frederike Lübeck and Roman Machacek and Theofilos Manitaras and Andreas Marfurt and Kyle Matoba and Simon Matrenok and Henrique Mendoncça and Fawzi Roberto Mohamed and Syrielle Montariol and Luca Mouchel and Sven Najem-Meyer and Jingwei Ni and Gennaro Oliva and Matteo Pagliardini and Elia Palme and Andrei Panferov and Léo Paoletti and Marco Passerini and Ivan Pavlov and Auguste Poiroux and Kaustubh Ponkshe and Nathan Ranchin and Javi Rando and Mathieu Sauser and Jakhongir Saydaliev and Muhammad Ali Sayfiddinov and Marian Schneider and Stefano Schuppli and Marco Scialanga and Andrei Semenov and Kumar Shridhar and Raghav Singhal and Anna Sotnikova and Alexander Sternfeld and Ayush Kumar Tarun and Paul Teiletche and Jannis Vamvas and Xiaozhe Yao and Hao Zhao Alexander Ilic and Ana Klimovic and Andreas Krause and Caglar Gulcehre and David Rosenthal and Elliott Ash and Florian Tramèr and Joost VandeVondele and Livio Veraldi and Martin Rajman and Thomas Schulthess and Torsten Hoefler and Antoine Bosselut and Martin Jaggi and Imanol Schlag},
      year={2025},
      eprint={2509.14233},
      archivePrefix={arXiv},
      primaryClass={cs.CL},
      url={https://arxiv.org/abs/2509.14233}, 
}

@inproceedings{vardial-2021,
  address =       {Kyiv, Ukraine},
  author =        {Chakravarthi, Bharathi Raja and Gaman, Mihaela and
                   Ionescu, Radu Tudor and Jauhiainen, Heidi and
                   Jauhiainen, Tommi and Lind{\'e}n, Krister and
                   Ljube{\v{s}}i{\'c}, Nikola and Partanen, Niko and
                   Priyadharshini, Ruba and Purschke, Christoph and
                   Rajagopal, Eswari and Scherrer, Yves and
                   Zampieri, Marcos},
  booktitle =     {Proceedings of the Eighth Workshop on NLP for Similar
                   Languages, Varieties and Dialects},
  editor =        {Zampieri, Marcos and Nakov, Preslav and
                   Ljube{\v{s}}i{\'c}, Nikola and Tiedemann, J{\"o}rg and
                   Scherrer, Yves and Jauhiainen, Tommi},
  month =         apr,
  pages =         {1--11},
  publisher =     {Association for Computational Linguistics},
  title =         {Findings of the {V}ar{D}ial Evaluation Campaign 2021},
  year =          {2021},
  url =           {https://aclanthology.org/2021.vardial-1.1/},
}

@inproceedings{vardial-2022,
  address =       {Gyeongju, Republic of Korea},
  author =        {Aepli, No{\"e}mi and Anastasopoulos, Antonios and
                   Chifu, Adrian-Gabriel and Domingues, William and
                   Faisal, Fahim and Gaman, Mihaela and
                   Ionescu, Radu Tudor and Scherrer, Yves},
  booktitle =     {Proceedings of the Ninth Workshop on NLP for Similar
                   Languages, Varieties and Dialects},
  editor =        {Scherrer, Yves and Jauhiainen, Tommi and
                   Ljube{\v{s}}i{\'c}, Nikola and Nakov, Preslav and
                   Tiedemann, J{\"o}rg and Zampieri, Marcos},
  month =         oct,
  pages =         {1--13},
  publisher =     {Association for Computational Linguistics},
  title =         {Findings of the {V}ar{D}ial Evaluation Campaign 2022},
  year =          {2022},
  url =           {https://aclanthology.org/2022.vardial-1.1/},
}

\clearpage
\onecolumn
\appendix

\section{Data Sources}
\label{app:data_sources}

\begin{table}[!ht]
\centering
\small
\begin{tblr}[
  label = {none},
  entry = {none},
]{
  colspec = {X[l] l r r r},
  hline{1} = {-}{},
  hline{2} = {-}{},
  hline{9,16,23,30,37,44} = {1}{},
  hline{9,16,23,30,37,44} = {2-5}{},
}
\parbox{4cm}{Source and URL to Dataset \\ (if available)} & Idiom    & Total \# of Samples & Total \# of Tokens & Avg. \# of Tokens/Sample \\
& Sursilvan  & 34,724              & 92,182          & 2.65                    \\
\SetCell[r=4]{l} \parbox{3cm}{Pledari Grond \\ \\ \url{https://www.pledarigrond.ch}} & Sutsilvan  & 32,175              & 307,799          & 9.57                    \\
& Surmiran & 43,954              & 301,390          & 6.86                    \\
& Puter    & 59,852              & 963,940          & 16.11                    \\
& Vallader & 72,470              & 1,141,688          & 15.75                    \\
& RG    & 172,616             & 785,831          & 4.55                     \\
& total    & 415,791             & 3,592,830         & 8.64                    \\
& Sursilvan  & 6,088               & 3,030,828             & 497.84                     \\
\SetCell[r=4]{l} \parbox{3cm}{La Quotidiana~(LQ) \\ \\  \url{https://huggingface.co/datasets/ZurichNLP/quotidiana}} & Sutsilvan  & 363               & 192,933             & 531.50                    \\
& Surmiran & 1,750               & 1,008,736             & 576.42                    \\
 & Puter    & 878               & 430,122             & 489.89                    \\
& Vallader & 2,395               & 1,250,042             & 521.94                    \\
& RG    & 2,567               & 1,365,988             & 532.13                    \\
& total    & 14,041               & 7,278,649            & 518.39                    \\
& Sursilvan  & 6,979               & 353,205            & 50.61                   \\
\SetCell[r=4]{l} \parbox{3cm}{Radiotelevisiun Svizra Rumantscha~(RTR) \\ \\ \url{https://developer.srgssr.ch/en/apis/rtr-linguistic}} & Sutsilvan  & 3,074                 & 156,674             & 50.97                   \\
 & Surmiran & 7,196                 & 245,931            & 34.18                   \\
& Puter    & 6,016                 & 225,389             & 37.46                   \\
& Vallader & 5,787               & 275,516          & 47.61                   \\
& RG    & 4,359               & 232,020          & 53.23                   \\
& total    & 33,411               & 1,488,735          & 44.56                   \\
& Sursilvan  & 4,931                & 641,844             & 130.17                    \\
\SetCell[r=4]{l} \parbox{3cm}{Telesguard Notes~(TG) \\ \\ n.a.}  & Sutsilvan  & 27                & 7,122             & 263.78                    \\
& Surmiran & 3,033                & 261,001             & 86.05                    \\
& Puter    & 578                & 162,754             & 281.58                     \\
& Vallader & 1,053                & 389,428             & 369.83                    \\
& RG    & 0                & 0             & 0.00                    \\
& total    & 9,622                & 1,462,149            & 151.96                    \\
& Sursilvan  & 12,233                & 971,643             & 79.43                    \\
\SetCell[r=4]{l} \parbox{3cm}{Mediomatix Textbooks (TB) \\ \\  \url{https://huggingface.co/datasets/ZurichNLP/mediomatix}} & Sutsilvan  & 12,116                & 1,005,356            & 82.98                    \\
& Surmiran & 6,698                & 534,563             & 79.81                    \\
& Puter    & 12,238                & 1,023,779            & 83.66                    \\
& Vallader & 12,283                & 1,019,339            & 82.99                    \\
& RG    & 0                   & 0                  & 0.00                        \\
& total    & 55,568                & 4,554,680            & 81.97                    \\

\end{tblr}
\caption{Summary of the number of samples extracted from each data source for each idiom, along with the number of whitespace tokens across all collected samples per data source and the average number of tokens per sample.}
\label{tab:data_splits_detailed}
\end{table}

\vfill
\pagebreak

\section{Data Split Statistics}
\label{app:data_stats}

\begin{table}[!ht]
\centering
\small
\begin{tblr}[
  label = {none},
  entry = {none},
]{
  colspec = {l l r r r},
  hline{1} = {-}{},
  hline{2} = {-}{},
  hline{9,16,23,30,37,44} = {1}{},
  hline{9,16,23,30,37,44} = {2-5}{},
}
Set       & Idiom    & Total \# of Samples & Total \# of Tokens & Avg. \# of Tokens/Sample \\
          & Sursilvan  & 55,574              & 3,663,641          & 65.92                    \\
          & Sutsilvan  & 44,188              & 1,342,065          & 30.37                    \\
          & Surmiran & 55,543              & 1,457,845          & 26.25                    \\
train-set & Puter    & 74,355              & 2,353,323          & 31.65                    \\
          & Vallader & 86,313              & 2,372,016          & 27.48                    \\
          & RG    & 171,199             & 1,190,808          & 6.96                     \\
          & total    & 487,172             & 12,379,698         & 25.41                    \\
          & Sursilvan  & 1,000               & 50,500             & 50.5                     \\
          & Sutsilvan  & 1,000               & 50,523             & 50.52                    \\
          & Surmiran & 1,000               & 35,133             & 35.13                    \\
dev-set   & Puter    & 1,000               & 36,891             & 36.89                    \\
          & Vallader & 1,000               & 48,482             & 48.48                    \\
          & RG    & 1,000               & 53,577             & 53.58                    \\
          & total    & 6,000               & 275,106            & 45.85                    \\
          & Sursilvan  & 1,000               & 502,709            & 502.71                   \\
          & Sutsilvan  & 100                 & 51,535             & 515.35                   \\
          & Surmiran & 800                 & 454,843            & 568.55                   \\
test-a    & Puter    & 100                 & 48,195             & 481.95                   \\
          & Vallader & 2,000               & 1,046,133          & 523.07                   \\
          & RG    & 2,000               & 1,067,778          & 533.89                   \\
          & total    & 6,000               & 3,171,193          & 528.53                   \\
          & Sursilvan  & 1,000                & 50,324             & 50.32                    \\
          & Sutsilvan  & 1,000                & 51,542             & 51.54                    \\
          & Surmiran & 1,000                & 33,906             & 33.91                    \\
test-b    & Puter    & 1,000                & 37,401             & 37.4                     \\
          & Vallader & 1,000                & 48,440             & 48.44                    \\
          & RG    & 1,000                & 52,523             & 52.52                    \\
          & total    & 6,000                & 274,136            & 45.69                    \\
          & Sursilvan  & 1,250                & 99,690             & 79.75                    \\
          & Sutsilvan  & 1,250                & 113,106            & 90.48                    \\
          & Surmiran & 1,000                & 79,629             & 79.63                    \\
test-c    & Puter    & 1,250                & 112,670            & 90.14                    \\
          & Vallader & 1,250                & 109,764            & 87.81                    \\
          & RG    & 0                   & 0                  & 0                        \\
          & total    & 6,000                & 514,859            & 85.81                    \\
          & Sursilvan  & 4,923                & 640,746            & 130.15                   \\
          & Sutsilvan  & 26                  & 7,120              & 273.85                   \\
          & Surmiran & 3,031                & 260,720            & 86.02                    \\
test-d    & Puter    & 576                 & 162,488            & 282.1                    \\
          & Vallader & 1,051                & 388,699            & 369.84                   \\
          & RG    & 0                   & 0                  & 0                        \\
          & total    & 9,607                & 1,459,773          & 151.95                   \\
\end{tblr}
\caption{Detailed statistics for all data splits by idiom, including number of samples, tokens, and average tokens per sample. The training set exhibits class imbalance, with RG containing the most samples but the shortest average length due to dictionary entries from Pledari Grond. \texttt{test-c} and \texttt{test-d} lack RG samples by construction.}
\label{tab:data_split_stats}
\end{table}

\end{document}